\documentclass[10pt,sigconf]{acmart}

\usepackage{amsmath,amsfonts}
\usepackage{graphicx}
\usepackage{textcomp}
\usepackage{xcolor}
\usepackage{algorithm, algpseudocode, xcolor, enumitem, float}
\usepackage[caption=false]{subfig}
\usepackage{varwidth}
\usepackage{caption}
\usepackage[framemethod=tikz]{mdframed}
\usepackage{hyperref}
\usepackage{dblfloatfix}
\usepackage{listings}
\usepackage{xcolor}
\usepackage{adjustbox}
\usepackage{array}

\newcolumntype{R}[2]{%
    >{\adjustbox{angle=#1,lap=\width-(#2)}\bgroup}%
    l%
    <{\egroup}%
}

\newcommand*\rot{\multicolumn{1}{R{50}{1em}}}

\usepackage{fancyhdr}

\copyrightyear{2023}
\acmYear{2023}
\setcopyright{licensedusgovmixed}\acmConference[ICPP-W 2023]{52nd
International Conference on Parallel Processing Workshops}{August 7--10,
2023}{Salt Lake City, UT, USA}
\acmBooktitle{52nd International Conference on Parallel Processing Workshops
(ICPP-W 2023), August 7--10, 2023, Salt Lake City, UT, USA}
\acmPrice{15.00}
\acmDOI{10.1145/3605731.3605886}
\acmISBN{979-8-4007-0842-8/23/08}

\begin{document}
\fancyhead{} 



\title{\huge Evaluation of OpenAI Codex for HPC Parallel Programming Models Kernel Generation}

\author{William F. Godoy, Pedro Valero-Lara, Keita Teranishi, Prasanna Balaprakash, and Jeffrey S. Vetter} 
\affiliation{ \institution{Oak Ridge National Laboratory\thanks{This manuscript has been authored by UT-Battelle LLC under contract DE-AC05-00OR22725 with the US Department of Energy (DOE). The publisher acknowledges the US government license to provide public access under the DOE Public Access Plan (\url{https://energy.gov/downloads/doe-public-access-plan}). "Godoy, 2023. This is the author's version of the work. It is posted here for your personal use. Not for redistribution. The definitive version was published in ICPP-W, \url{https://doi.org/10.1145/3605731.3605886}."
}} 
\city{Oak Ridge}
\state{TN}
\country{USA} }
\email{{godoywf}, {valerolarap},{teranishik},{pbalapra},and {vetter}@ornl.gov}



\begin{abstract}
We evaluate AI-assisted generative capabilities 
on fundamental numerical kernels in high-performance computing (HPC), including AXPY, GEMV, GEMM, SpMV, Jacobi Stencil, and CG. We test the generated kernel codes for a variety of language-supported programming models, including (1) C\texttt{++} (e.g., OpenMP [including offload], OpenACC, Kokkos, SyCL, CUDA, and HIP), (2) Fortran (e.g., OpenMP [including offload] and OpenACC), (3) Python (e.g., numba, Numba, cuPy, and pyCUDA), and (4) Julia (e.g., Threads, CUDA.jl, AMDGPU.jl, and KernelAbstractions.jl). 
We use the GitHub Copilot capabilities powered by OpenAI Codex available in Visual Studio Code as of April 2023 to generate a vast amount of implementations given simple \texttt{<kernel> + <programming model> + <optional hints>} prompt variants. To quantify and compare the results, we propose a proficiency metric around the initial 10 suggestions given for each prompt. 
Results suggest that the OpenAI Codex outputs for C\texttt{++} correlate with the adoption and maturity of programming models. For example, OpenMP and CUDA score really high, whereas HIP is still lacking. 
We found that prompts from either a targeted language such as Fortran or the more general-purpose Python can benefit from adding code keywords, while Julia prompts perform acceptably well for its mature programming models (e.g., Threads and CUDA.jl). 
We expect for these benchmarks to provide a point of reference for each programming model's community.
Overall, understanding the convergence of large language models, AI, and HPC is crucial due to its rapidly evolving nature and how it is redefining human-computer interactions.
\end{abstract}



\maketitle
\newcommand{\addtab}[3]{
    \begin{table}[hbt!]
        \centering
        \caption{\footnotesize{#2}}
        \label{#1}
        \includegraphics[width=\linewidth]{#3}
    \end{table}
}
\newcommand{\addfigwidth}[4]{
\begin{figure}[hbt!]
  \centering
  \includegraphics[width=#4\textwidth]{#3}
  \vspace{-0.05in}
  \caption{\footnotesize{#2}}
  \label{#1}
  \vspace{-0.15in}
\end{figure}
}
\newcommand{\addfig}[3]{
    \begin{figure}[hbt!]
        \centering
        \includegraphics[width=\linewidth,height=\textheight,keepaspectratio]{#3}
        \caption{\footnotesize{#2}}
        \label{#1}
    \end{figure}
}
\newcommand{\addmediumfig}[3]{
    \begin{figure}[hbt!]
        \centering
        \includegraphics[width=3.5in]{#3}
        \caption{\footnotesize{#2}}
        \label{#1}
    \end{figure}
}
\newcommand{\addverysmallfig}[3]{
    \begin{figure}[hbt!]
        \centering
        \includegraphics[width=1.8in]{#3}
        \caption{\footnotesize{#2}}
        \label{#1}
    \end{figure}
}
\newcommand{\addsmallfig}[3]{
    \begin{figure}[hbt!]
        \centering
        \includegraphics[width=2.5in]{#3}
        \caption{\footnotesize{#2}}
        \label{#1}
    \end{figure}
}
\newcommand{\addsmallalgo}[4]{
    \begin{algorithm}[hbt!]
        \centering
        \includegraphics[width=#4in]{#3}
        \caption{\footnotesize{#2}}
        \label{#1}
    \end{algorithm}
}
\newcommand{\addsubcorefig}[5]{
\begin{subfigure}[b]{#4\textwidth}
  \includegraphics[width=2in, height=2in]{#3}
  \vspace{#5}
  \caption{\footnotesize{#2}}
  \label{#1}
\end{subfigure}
}
\newcommand{\addtwofig}[8]{
\begin{figure}[hbt!]
  \centering
  \addsubcorefig{#3}{#4}{#5}{0.3}{-0.20in}
  \addsubcorefig{#6}{#7}{#8}{0.3}{-0.20in}
  \caption{\footnotesize{#2}}
  \label{#1}
\end{figure}
}
\newcommand{\addxilinxdesignfigcore}[5]{
\begin{subfigure}[b]{#4\textwidth}
  \includegraphics[width=1.7in, height=0.8in]{#3}
  \vspace{#5}
  \caption{\footnotesize{#2}}
  \label{#1}
\end{subfigure}
}
\newcommand{\addxilinxdesignfig}[8]{
\begin{figure}[hbt!]
  \addxilinxdesignfigcore{#3}{#4}{#5}{0.23}{-0.20in}
  \addxilinxdesignfigcore{#6}{#7}{#8}{0.23}{-0.20in}
  \vspace{-0.32in}
  \caption{\footnotesize{#2}}
  \label{#1}
  \vspace{-0.15in}
\end{figure}
}
\newcommand{\addexamplefigcore}[5]{
\begin{subfigure}[b]{#4\textwidth}
  \includegraphics[width=1.7in, height=1.4in]{#3}
  \vspace{#5}
  \caption{\footnotesize{#2}}
  \label{#1}
\end{subfigure}
}
\newcommand{\addexamplefig}[8]{
\begin{figure}[hbt!]
  \addexamplefigcore{#3}{#4}{#5}{0.4}{-0.20in}
  \addexamplefigcore{#6}{#7}{#8}{0.4}{-0.20in}
  \caption{#2}
  \label{#1}
\end{figure}
}
\newcommand{\addalgo}[3]{
    \begin{algorithm}[hbt!]
        \centering
        \includegraphics[width=\linewidth,height=\textheight,keepaspectratio]{#3}
        \caption{\footnotesize{#2}}
        \label{#1}
    \end{algorithm}
}
\def\algoref#1{Algorithm~\ref{#1}}
\def\figref#1{Figure~\ref{#1}}
\def\tabref#1{Table~\ref{#1}}
\def\eqref#1{Equation~\ref{#1}}
\newcommand{\specialcell}[2][c]{%
  \begin{tabular}[#1]{@{}c@{}}#2\end{tabular}}

%
%

\section{Introduction}
\label{sec:Introduction}

Since its initial release in 2020, the Generative Pre-trained Transformer 3 (GPT-3)~\cite{NEURIPS2020_1457c0d6} has signified a revolutionary step in the evolution of human-computer interactions. Developed by OpenAI,\footnote{\url{https://openai.com/}} GPT-3 is the third-generation prediction-based large language model (LLM) used for several AI-generated, human-like text applications. It has received praise for the high-quality results in several natural language processing (NLP) tasks~\cite{doi:10.1126/science.aaa8685} due in part to the unprecedented investment (\$12 million USD) and size of its training model (175 billion parameters at 800\,GB). Hence GPT-3 and its successor GPT-4\footnote{\url{https://openai.com/product/gpt-4}} are defining several societal questions for the near future.

As we enter the exascale computing era, which is dominated by the extreme heterogeneity of hardware and programming models~\cite{vetter:2018:extreme}, AI-assisted code generation could play a key role in how we develop, deploy, and test software that targets high-performance computing (HPC) systems. Traditional human-readable code such as C\texttt{++}~\cite{stroustrup2013c++}, Fortran~\cite{Fortran}, Python~\cite{Python}, and more recently Julia~\cite{Bezanson2017-ca}, are a straightforward application for GPT-3 LLM capabilities that would help redefine software development. Hence, we need to understand the current state of practice, limitations, and potential of this new technology.

Our early experiences with OpenAI Codex, which is a GPT-3 descendant, came from using the GitHub Copilot\footnote{\url{https://github.com/features/copilot}} plugin available on Visual Studio Code\footnote{\url{https://docs.github.com/en/copilot/getting-started-with-github-copilot?tool=vscode}} to generate relevant scientific kernels for HPC.
Our goal is to assess and understand the impact of GPT-3's state-of-art capabilities in the overall interactive process for generating, optimizing, building, and testing well-established mathematical HPC kernels. For our assessment, we evaluate AXPY, general matrix-vector multiply (GEMV), general matrix-matrix multiply (GEMM), sparse matrix–vector multiplication (SpMV), 3D Jacobi stencil computations, and conjugate gradients (CG). We test the generation of these kernels over a variety of programming models that target CPU and GPU hardware. Our goal is to establish a benchmark and understand the status of \textit{prompt} engineering in the OpenAI Codex when applied to these important kernels and programming models as the technology continues to evolve. 

The rest of the paper is structured as follows: Section~\ref{sec:Background} provides an overview of related efforts that highlight the recent attention to these topics in the broader area of computer science. 
We describe our prompt input pattern methodology for interacting  with GitHub Copilot for the kernel generation process along with the proposed metric to score the quality of the suggested outputs in Section~\ref{sec:Methodology}. Section~\ref{sec:Results} presents the results of our evaluation along with our findings on prompt trade-off options for each language, kernel, programming model, and additional keyword inputs on the generated outputs. We aim to understand the current status of the correctness, trade-offs, and overall value of LLMs for HPC practitioners who wish to adopt this technology.
Conclusions and future directions are presented in Section~\ref{sec:Conclusions}.
Appendix~\ref{ap1:Artifact} provides the artifact description for the reproducibility of this study.

\section{Background}
\label{sec:Background}

The availability of GPT-3 has led to increasing interest in its LLM capabilities and how they could be applied to a large variety of applications. 
Brown et al.~\cite{NEURIPS2020_1457c0d6} introduced the seminal paper on GPT-3's unprecedented results, which showed strong performance in NLP interactions such as translation, question-answering, and cloze tasks. They highlight the use and success rates of prompt-based~\cite{NIPS2004_ef1e491a,1597116} zero, one, and few-shot learning (FSL) techniques when only a few data examples are available to train a model as opposed to other previously trained ML alternatives. Wang et al.~\cite{10.1145/3386252} presented a comprehensive review of FSL and pointed out the ``unreliable empirical risk minimizer'' that, due to its nature, compensates for the lack of supervised information by using prior available knowledge. Floridi and Chiriatti~\cite{floridi2020gpt} provide a commentary on the nature of GPT-3, its scope, and limits while outlining the social consequences of the ``industrialization of automatic cheap production of good, semantic artifacts.''

On the code generation side, Dehaerne et al.~\cite{9849664} provided a systematic review of applying ML methods in description-to-code, code-to-code, and code-to-description studies from the past 6 years. The review indicated limitations, including the variability in the quality of generated, mined source code and the need and availability of large datasets (e.g., GitHub sources). The review also highlighted that automatically generating data is a fast alternative for obtaining data but is only appropriate for certain contexts, such as \textit{programming by example}. Recent work has also focused on using GitHub Copilot's \textit{AI pair programmer}, which is based on OpenAI Codex and leverages the vast stores of source code hosted on GitHub for AI-assisted code generation. 
Chen et al.~\cite{chen2021evaluating} provided an introduction and evaluation of Codex for its Python code-writing capabilities.  They point out the current limitations and difficulty with ``docstrings describing long chains of operations and with binding operations to variables.'' A major challenge noted is the overreliance on Codex by novice programmers.
Importantly, Copilot uses a different Codex version than the one used in our study.
Nguyen and Nadi~\cite{10.1145/3524842.3528470} provided an empirical evaluation of GitHub Copilot suggestions for correctness and understandability to LeetCode questions in Java, JavaScript, Python, and C. Shortcomings included generating further simplified code that relies on undefined functions. Vaithilingam, Zhang, and Glassman~\cite{10.1145/3491101.3519665} provided a human subject study that consisted of 24 participants on the usability of GitHub Copilot. They concluded that Copilot did not necessarily reduce the task completion time in common tasks, such as file editing, web scraping, and graph plotting, for experienced Python users. Sobania et al.~\cite{10.1145/3512290.3528700} found little difference between Copilot and automatic generators that use genetic programming when applied to the PSB2 program synthesis benchmarks~\cite{10.1145/3449639.3459285}. 
Imai~\cite{10.1145/3510454.3522684} provided a preliminary assessment that showed that, although Copilot helps generate lines of code faster, the quality is lower when compared to human pair programming. Yetistiren et al.~\cite{10.1145/3558489.3559072} assessed the correctness, validity, and efficiency of targeting the HumanEval problem dataset~\cite{chen2021evaluating} with a high success rate.

From an educational perspective, Sarsa et al.~\cite{10.1145/3501385.3543957} discussed Codex's effectiveness in generating programming exercises and explanations via Copilot, not without addressing the need for oversight in output quality. Finnie-Ansley et al.~\cite{10.1145/3511861.3511863} reported that Codex ranks high when compared to a class of introductory computer science students taking programming tests. Similarly, Denny et al.~\cite{10.1145/3545945.3569823} discussed the pedagogical value of Copilot for testing their answers and for its prompt nature in introductory programming questions. 
Wermelinger~\cite{10.1145/3545945.3569830} discussed important questions about how teaching programming will evolve.
Similarly, Brennan and Lesage~\cite{10.1007/978-3-031-24291-5_20} discussed the educational opportunities for undergraduate engineering and the need for students to have a strong intuition for software development when using AI-assisted tools.

Pierce et al.~\cite{9833571} assessed security aspects of Copilot's code contributions in a large sample, and they found relatively high vulnerability rates. They also concluded that there is a need to quantify the limits of generated code in low-level scenarios, and Copilot must be paired with security-aware tooling. 

Overall, there is a common thread in the relevant literature: the GPT-based tools are here to stay and will impact almost every aspect of human-computer interaction. However, we are at an inflection point where more exploratory research is needed when assessing AI code-generating capabilities, and their technical, economic, and social implications must be carefully explored and better understood. To the best of our knowledge there is no reported work on the application of current Copilot capabilities for HPC kernels.

\section{Methodology}
\label{sec:Methodology}

We evaluate prompt outputs for parallel programming models available in four different languages: C\texttt{++}, Fortran, Python, and Julia.
Our methodology has two aspects: (1) the Copilot code suggestion generation from prompt queries based on kernels and the parallel programming model for each language and (2) a simple metric to evaluate the correctness of the results. As stated in their documentation,\footnote{\url{https://docs.github.com/en/copilot/overview-of-github-copilot/about-github-copilot-for-individuals}} GitHub Copilot is ``trained on all languages that appear in public repositories. 
For each language, the quality of suggestions you receive may depend on the volume and diversity of training data for that language.'' Hence, we investigate whether or not these results correlate with the expected availability of correct programming models and public code examples. We will use assess popularity, in particular for open-source software~\cite{9434501},  
based on (1) the number of repositories per language on GitHub (GitHut\footnote{\url{https://madnight.github.io/githut/\#/pull_requests/2023/1}}) 
and (2) the TIOBE index.\footnote{\url{https://www.tiobe.com/tiobe-index/}} 
Hence, C\texttt{++} and Python codes are expected to have wider availability than Julia and Fortran. Nevertheless, because of the general-purpose nature of C\texttt{++} and Python, this metric might not be as relevant as it is for scientific and mathematical applications written in Fortran or Julia.

We assess the outputs from each prompt and categorize based on their level of correctness. The rest of the section describes the experiment setup and the proposed metric to quantify the quality of the AI-generated results.

\subsection{Experiment setup}
We select a simple prompt base structure for each targeted language. Visual Studio Code recognizes the targeted language through the appropriate support plugin and the currently opened file based on the language extension (e.g., cpp, f90, py, jl) and makes it part of the prompt prefix.
The rest of the prompt is generated through a comment line complemented with optional code. Pressing \texttt{<Control> + <Enter>} results in a separate window that contains a maximum of 10 code suggestions.  Given the highly sensitive nature of the LLM, even one additional keyword could heavily influence the output. 
 
\begin{figure}[!h]
\subfloat[]{\includegraphics[scale=0.23]{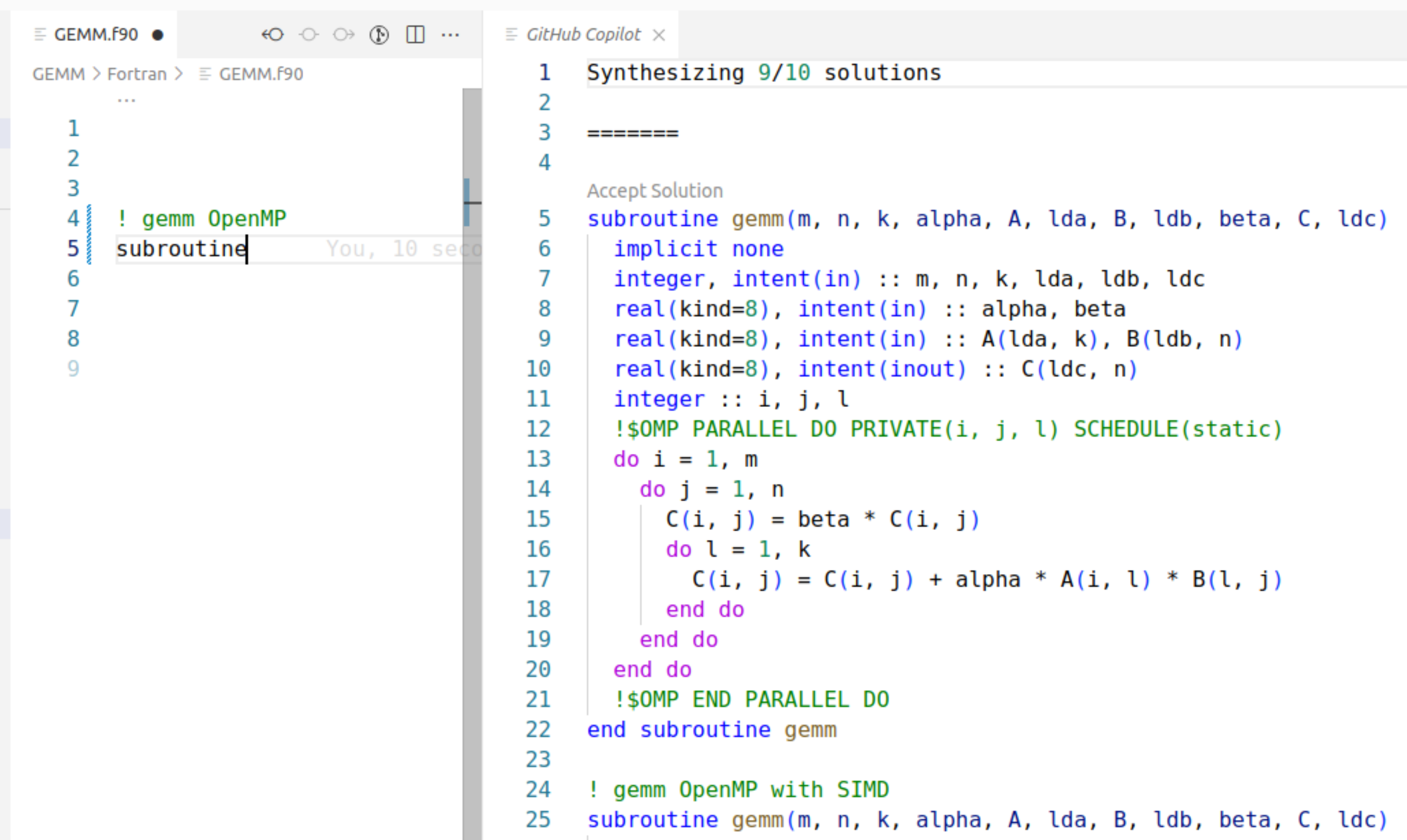}
\label{fig:copilot-fortran}}
\hfil
\subfloat[]{\includegraphics[scale=0.26]{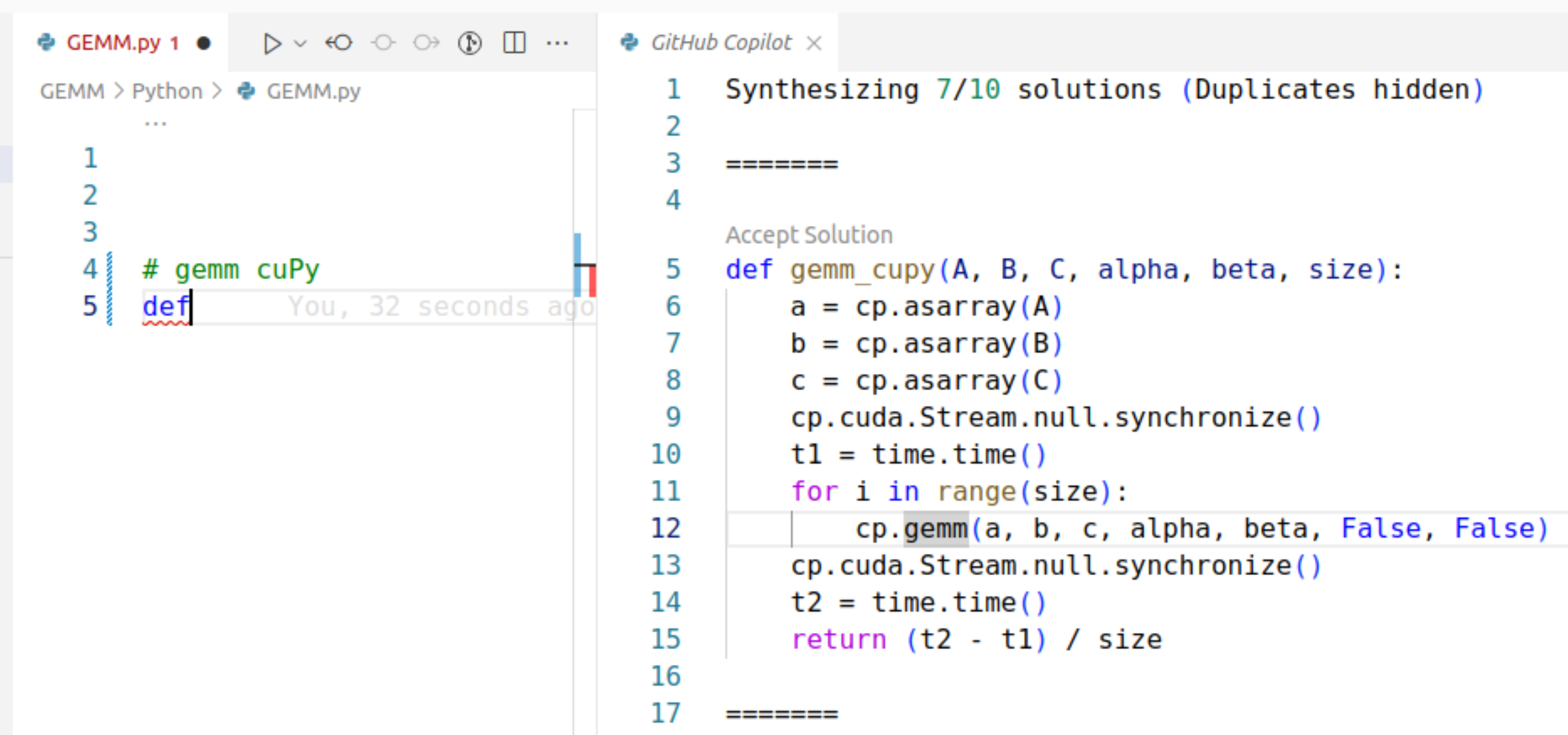}
\label{fig:copilot-python}}
\caption{GitHub Copilot prompt interface with optional language keywords for (a) Fortran's \texttt{subroutine} and (b) Python's \texttt{def}.}
\end{figure}

Therefore, our simple prompt queries can be described by using the following structure:

\begin{itemize}
    \item <kernel> <programming-model>
    \item <kernel> <programming-model> <optional keywords: \texttt{function}, \texttt{subroutine}, and \texttt{def}>
\end{itemize}

Table~\ref{tab:scope} shows the programming language and model combinations used for this study. The additional post-fix optional keywords correspond to our attempt to add more information to the prompt query to obtain better-quality suggestions. As shown in Section~\ref{sec:Results}, C\texttt{++} model sensitivity can vary by using the word \textit{function} (not a language keyword), whereas
Fortran and Python are consistently sensitive to the \texttt{subroutine} and \texttt{def} keywords. We also found little sensitivity in Julia prompts when adding a post fix (e.g., \texttt{function}), hence we did not add it in this study.

\begin{table}[!ht]
\begin{center}
\begin{tabular}{l l l }
 Language    & Programming model & Post fix \\
 \hline
 C\texttt{++} & OpenMP~\cite{openmp}           & offload, function \\
              & OpenACC~\cite{openacc}         & function \\
              & Kokkos~\cite{Kokkos}           & function \\
              & CUDA~\cite{cuda}             & function \\
              & HIP~\cite{hip}              & function \\
              & Thrust~\cite{Thrust}           & function \\
 \hline
 Fortran      & OpenMP           & offload, subroutine \\
              & OpenACC          & subroutine \\
 \hline
 Python       & numpy~\cite{van2011numpy}            & def \\
              & Numba~\cite{lam2015numba}            & def \\
              & pyCUDA~\cite{KLOCKNER2012157}           & def \\
              & cuPy~\cite{nishino2017cupy}             & def \\
\hline
Julia         & Threads~\cite{knopp2014experimental}    & \\
              & CUDA~\cite{besard2018juliagpu}          & \\
              & AMDGPU~\cite{AMDGPU}           & \\
              & KernelAbstractions~\cite{KA} & \\
\hline
\hline
 \end{tabular}
 \end{center}
 \caption{Scope of our experimental setup applied for each kernel in terms of language and targeted parallel programming model.}
 \label{tab:scope}
 \end{table}
\vspace{-1cm}
\subsection{Correctness metric}
To evaluate the correctness of the generated suggestions, we propose a simple approach that is based on our observations more than on any particular formalism.
We consider five levels of correctness and proficiency labels between [0], or {\it non-knowledge},  and [1], or {\it expert}, when observing the suggested answers given by Copilot as those illustrated in Figures~\ref{fig:copilot-fortran} and~\ref{fig:copilot-python} for each combination in Table~\ref{tab:scope}.

\begin{itemize}
    \item [0] {\it non-knowledge}: No code at all or not a single correct code.
    \item [0.25] {\it novice}: One correct code, but it includes other several correct or incorrect programming models (e.g., OpenACC suggestions in an OpenMP prompt).
    \item [0.5] {\it learner}: One correct code, and there are other incorrect codes, but all of them are using the requested programming model.
    \item [0.75] {\it proficient}: All codes are correct and using the programming model requested.
    \item [1] {\it expert}: Only one piece of code is provided, and it is totally correct.
\end{itemize}

\section{Results}
\label{sec:Results}

We present our analysis for each selected language and kernel and discuss 
some observations from our initial interaction with GitHub Copilot.
As shown in Figures~\ref{fig:copilot-fortran} and \ref{fig:copilot-python}, for each combination listed in Table\ref{tab:scope}, we obtain a set of suggestions for each prompt. This approach enables us to evaluate the effectiveness of OpenAI Codex in generating accurate and proficient code. Notably, none of the experiments reached a perfect score of 1, which illustrates the stochastic status of the current LLM when applied to a particular area.

\subsection{C++}

\begin{figure}[!h]
\begin{center}
\includegraphics[width=0.45\textwidth]{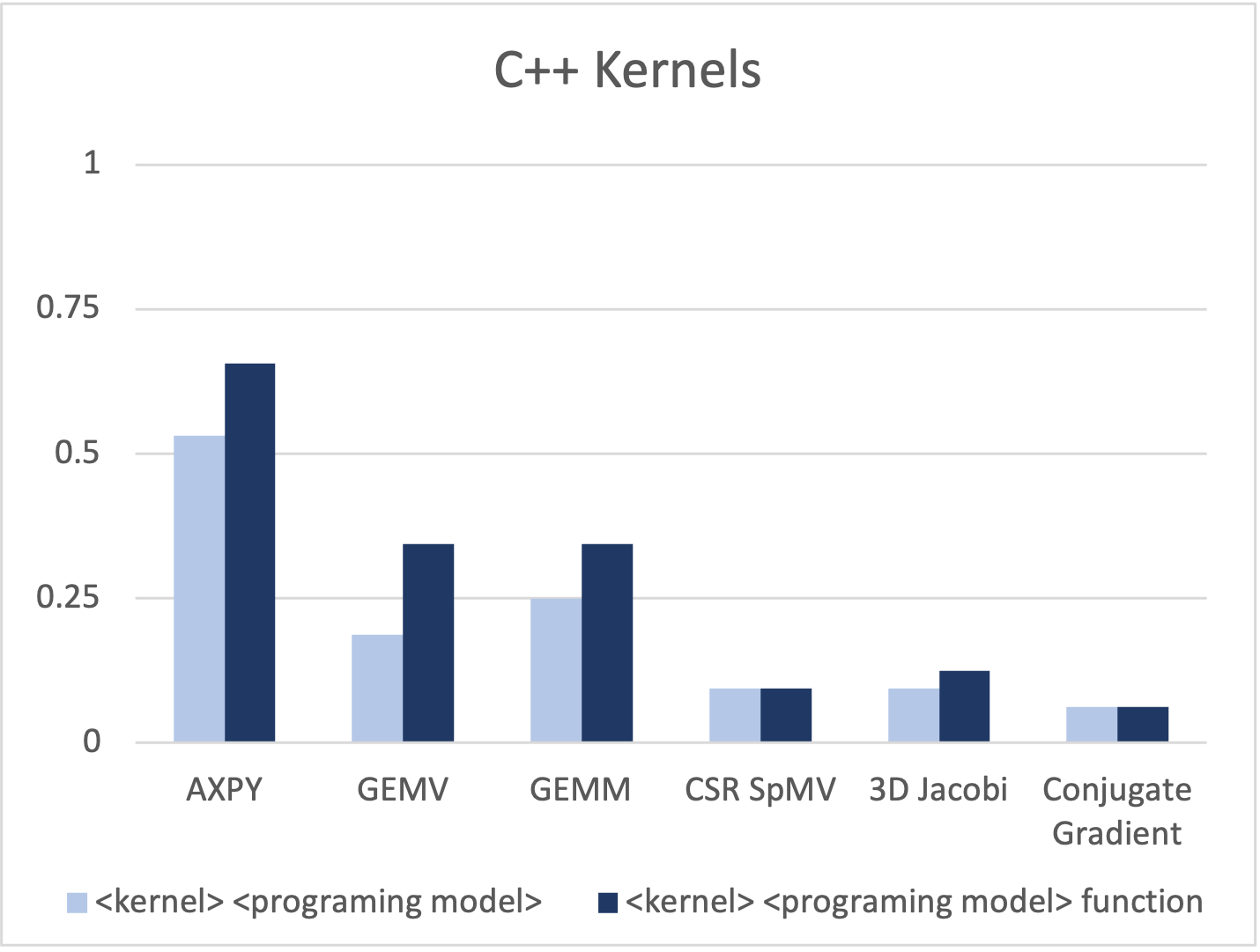}
\includegraphics[width=0.45\textwidth]{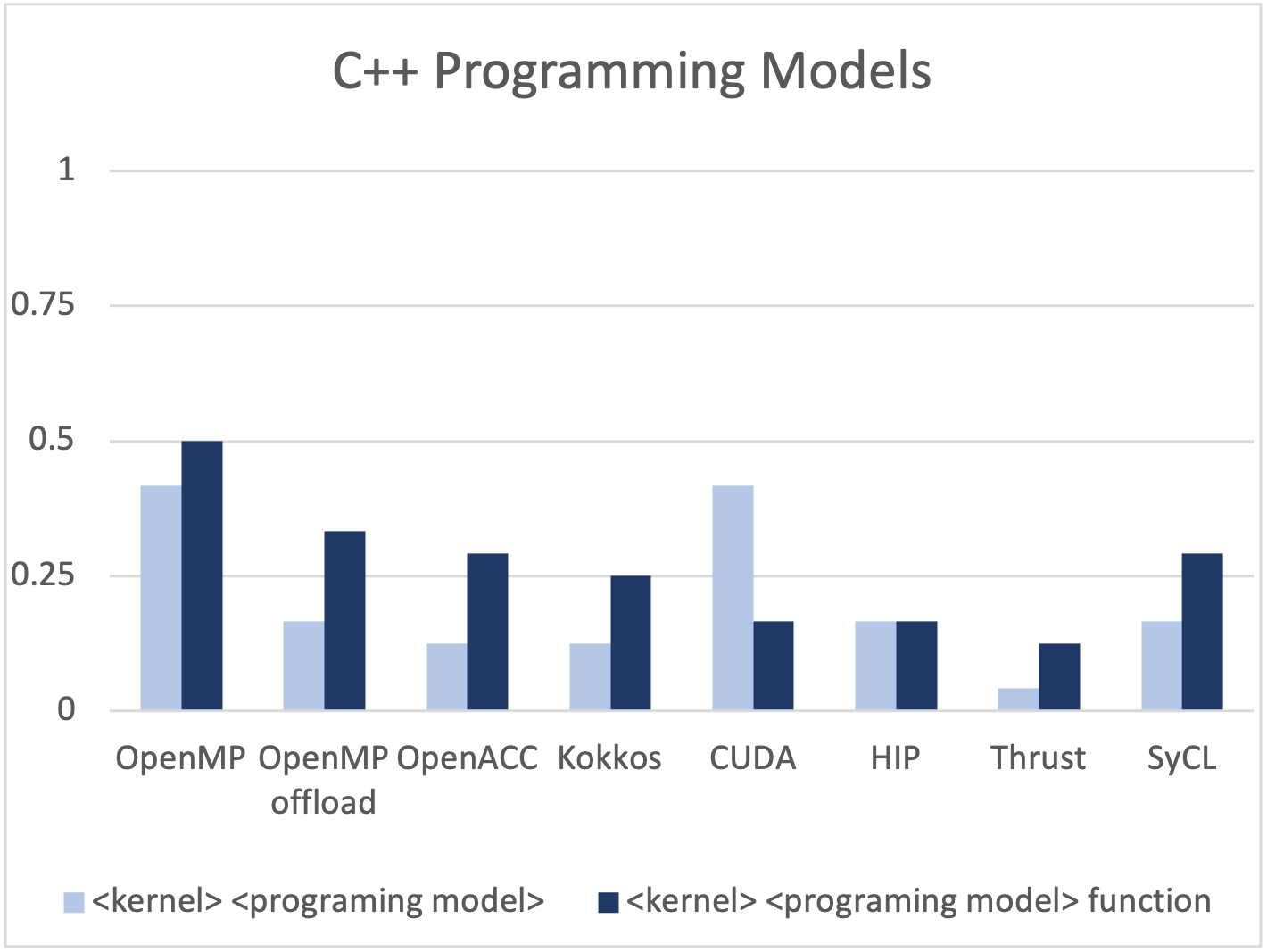}
\caption{Results for C++ kernels (top) and programming models (bottom).}
\label{fig:c++results}
\end{center}
\end{figure}

Table~\ref{tab:results-cpp} shows the results for all of our C\texttt{++} experiments, and Figure~\ref{fig:c++results} illustrates the results according to the different kernels and programming models.

\begin{table}[h]
\begin{center}
\begin{tabular}{lllllll}
 Prompt         &  \rot{AXPY} & \rot{GEMV} & \rot{GEMM} & \rot{SpMV} & \rot{Jacobi} & \rot{CG} \\        
 \hline
Prefix <kernel>   &        &       &        &         &         &      \\
OpenMP            &  0.75  &  0.5  &  0.5   &  0.5    &  0      &  0.25 \\
OpenMP offload    &  0.5   &  0.5  &  0.5   &  0.25   &  0.25   &  0   \\
OpenACC           &  0.5   &  0    &  0.25  &  0      &  0      &  0   \\
Kokkos            &  0.5   &  0    &  0     &  0      &  0.25   &  0   \\
CUDA              &  0.75  &  0.75 &  0.75  &  0      &  0      &  0.25 \\
HIP               &  0.75  &  0    &  0     &  0      &  0.25   &  0   \\
Thrust            &  0.25  &  0    &  0     &  0      &  0      &  0   \\
SyCL              &  0.75  &  0.25 &  0     &  0      &  0      &  0   \\

\hline
Prefix <kernel>        &       &      &      &      &      &     \\
Post fix function  &       &      &      &      &      &     \\
OpenMP               &  0.75  & 0.75  & 0.75  & 0.25  & 0.25   & 0.25 \\
OpenMP offload       &  0.5   & 0.5   & 0.5   & 0.25  & 0.25   & 0   \\
OpenACC              &  0.5   & 0.5   & 0.5   & 0.25  & 0  & 0  \\
Kokkos               &  0.75  & 0.25  & 0.25  & 0     & 0.25   & 0   \\
CUDA                 &  0.75  & 0.25  & 0     & 0     & 0   & 0   \\
HIP                  &  0.75  & 0     & 0     & 0     & 0.25   & 0   \\
Thrust               &  0.5   & 0     & 0.25  & 0     & 0  & 0   \\
SyCL                 &  0.75  & 0.5   & 0.25  & 0     & 0  & 0   \\
\hline
\hline
\end{tabular}
\caption{Metric assessment for GitHub Copilot's C\texttt{++} outputs using the input prompt pattern \texttt{<kernel> <programming model> (function)}.}
\label{tab:results-cpp}
\end{center}
\end{table}

\vspace{-0.75cm}

As shown, the best results are achieved for the AXPY kernel. We see a clear trend in these results (Figure~\ref{fig:c++results} [left]): the more complex the kernel, the fewer quality results are obtained. 
Although we see a level between {\it learner} and {\it proficient} for the AXPY kernel (the simplest kernel), for CG (the most complicated kernel), the level is close to {\it non-knowledge}.
For the programming models, we see better results for OpenMP and CUDA. This could be due to the maturity of these programming models compared to others and their availability in public code. We note that the use of \texttt{function} as part of the prompt is very beneficial and increments the quality of the results in the dense matrix cases. Sparse matrices and high-level algorithms (e.g., SpMV, Jacobi, and CG) do not show much improvement. 
However, a different trend is observed for CUDA, for which, instead of improving the level of proficiency, it actually decreases the quality of the answer. This is perhaps because the word \texttt{function} is not used for CUDA codes. In our experimentation, which is not shown here, using the words ``kernel'' or ``\texttt{\_\_global\_\_}'' led to better code generation quality. This is an example of how important it is to adapt the language used for the prompts to the particularities or syntax habitually used by such a community.  

Results also show that high-level programming model prompts from Kokkos, Thrust, and SyCL perform poorly over several kernels. We view these results as a reflection of the user community size of these high-level abstractions. Over several instances, we observed that many wrong answers or no answers at all dominate as the kernel becomes more complex. Notably, large benchmark repositories are available (e.g., HecBench~\cite{jin2023hecbench,jin2021rodinia}), from which some of the responses originate.

\subsection{Fortran} 

Fortran is of particular interest in this analysis owing to its importance in  HPC and scientific computing. Despite not being a mainstream language in terms of code availability, Copilot can provide some good results because of Fortran's domain-specific nature and legacy.

As shown in Table~\ref{tab:fortranresults}, using an ``optimized'' prompt and the \texttt{subroutine} keyword is particularly beneficial in this case. Not using it leads to very poor results, with the AXPY OpenMP case being the only exception due to its simplicity and availability.
We observe a trend similar to the one we saw in the C\texttt{++} case: the more mature solutions such as OpenMP and OpenACC provide better results for parallel codes that use Fortran. 

\begin{table}[!h]
\begin{center}
\begin{tabular}{lllllll}
 Prompt         &  \rot{AXPY} & \rot{GEMV} & \rot{GEMM} & \rot{SpMV} & \rot{Jacobi} & \rot{CG} \\        
 \hline
Prefix <kernel>   &       &      &     &       &      &      \\
OpenMP            &  0.75  &  0   &  0  &  0  &  0   &  0   \\
OpenMP offload    &  0    &  0   &  0  &  0    &  0   &  0   \\
OpenACC           &  0    &  0   &  0  &  0  &  0   &  0   \\
\hline
Prefix <kernel>        &       &      &      &      &      &     \\
Post fix subroutine  &       &      &      &      &      &     \\
OpenMP               &  0.75  & 0.25  & 0.25  & 0.5   & 0.5   & 0.25   \\
OpenMP offload       &  0.25  & 0.25  & 0.25  & 0.25  & 0.5   & 0.25   \\
OpenACC              &  0.25  & 0.25  & 0.25  & 0.25  & 0.25  & 0.25   \\
\hline
\hline
\end{tabular}
\caption{Metric assessment for Copilot's outputs using the input prompt pattern \texttt{<kernel> <programming model> (subroutine)} for Fortran.}
\label{tab:fortranresults}
\end{center}
\end{table}

\begin{figure}[!h]
\begin{center}
\includegraphics[width=0.45\textwidth]{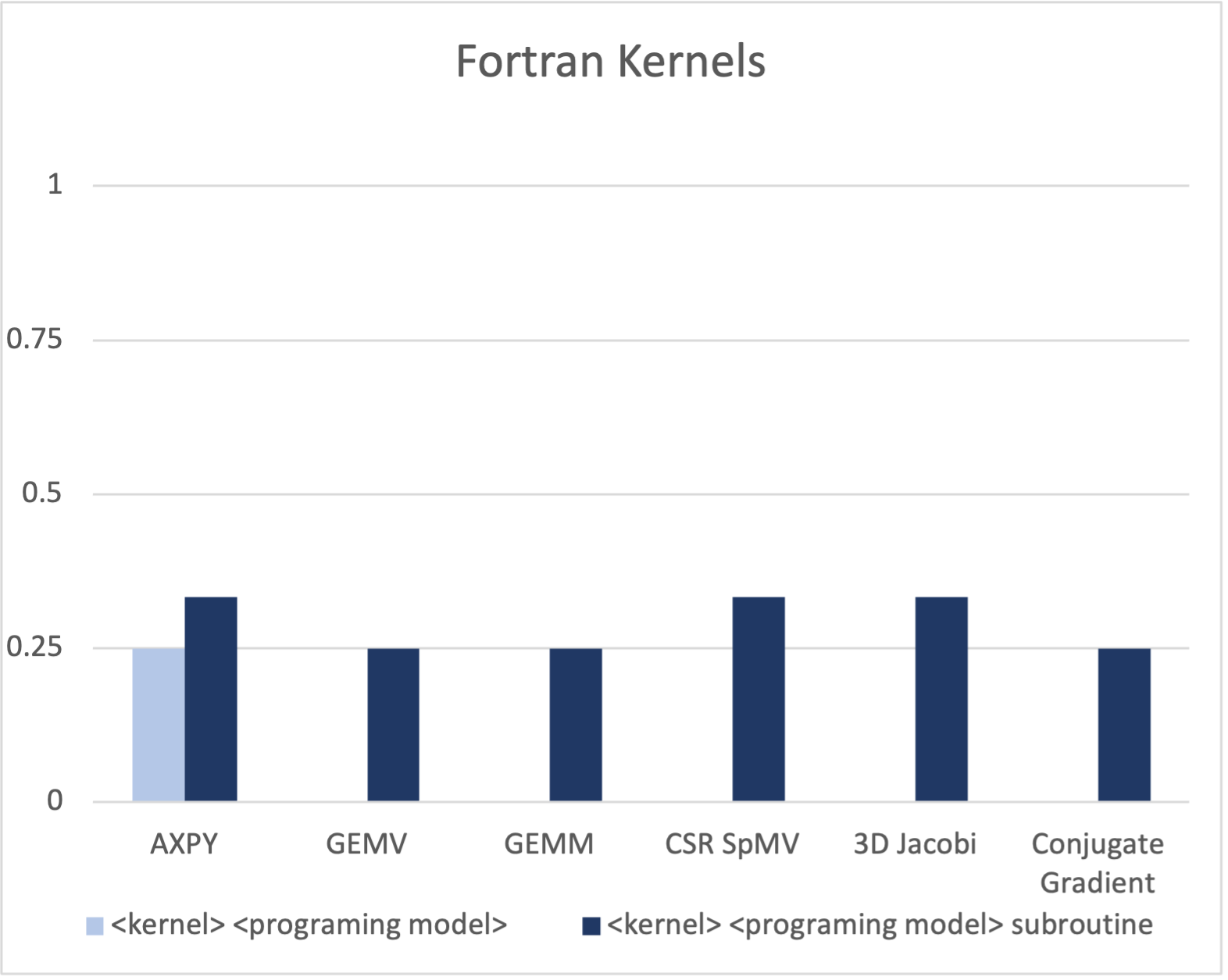}
\includegraphics[width=0.45\textwidth]{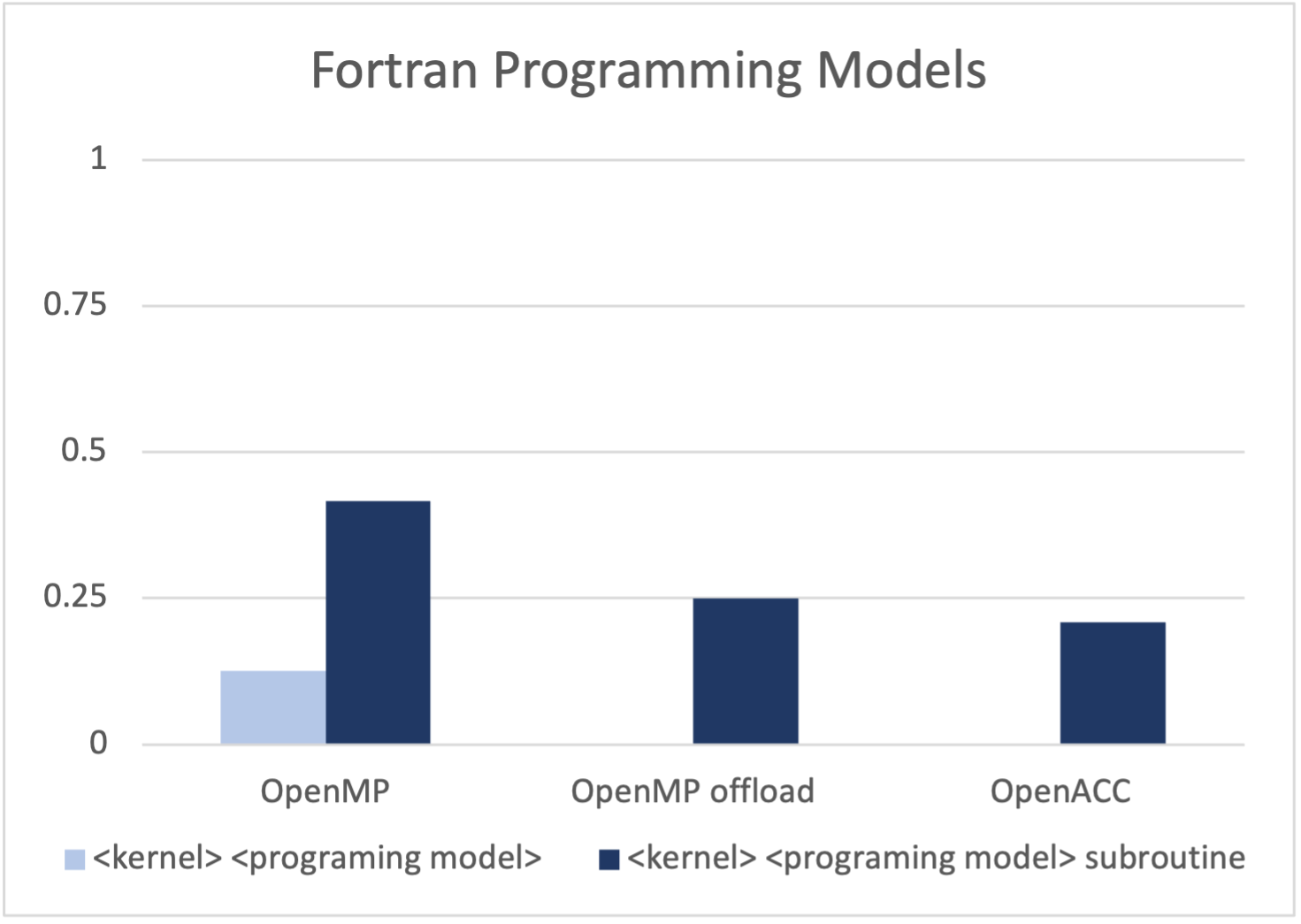}
\caption{Results for Fortran kernels (top) and programming models (bottom).}
\label{fig:fortranresults}
\end{center}
\end{figure}

\vspace{-0.75cm}
Figure~\ref{fig:fortranresults} illustrates the uniform characteristics in the responses across kernels. Fortran is used in many legacy HPC applications, and with so many available Fortran codes to draw from, it is not surprising that Copilot can obtain correct kernel implementations with higher complexity kernels (e.g., CG, Jacobi) in addition to the simpler ones (e.g., AXPY, GEMM).

\subsection{Python}
Python is one of the most-used, general-purpose languages in industry, research, and AI and also has an important role in education and as a major target of AI-generative code, as documented in Section~\ref{sec:Background}.
Although neither a parallel programming model nor part of the standard,  we consider \texttt{numpy} in our evaluation because it is the de-facto standard for scientific computing in Python.

Table~\ref{tab:pythonresults} shows the results from our evaluation. Similar to Fortran, we observe that the quality improves dramatically when adding the Python \texttt{def} keyword to clarify that we are looking for functions in the language. Overall, Copilot can generate acceptable numpy, cuPy, and pyCUDA implementations across several kernels, but Numba falls behind. Perhaps this is an indication that cuPy and pyCUDA are more popular alternatives in the community via lightweight layers on top of compiled C or CUDA code, rather than using Numba's just-in-time approach on top of LLVM. Notably, Numba recently deprecated support for AMD GPU hardware, hence reinforcing the idea that GPU kernels written in Python favor CUDA-like implementations. Interestingly, successful GPU instances (e.g., pyCUDA and cuPy) include a correct raw CUDA kernel source code as a user-defined kernel. These instances are described in  pyCUDA~\cite{kloeckner22pycuda} and cuPy~\cite{CuPyDocument} documentation, respectively.

\begin{table}[!h]
\begin{center}
\begin{tabular}{lllllll}
 Prompt         &  \rot{AXPY} & \rot{GEMV} & \rot{GEMM} & \rot{SpMV} & \rot{Jacobi} & \rot{CG} \\        
 \hline
Prefix <kernel>   &       &      &       &      &      &      \\
numpy             &  0.25  &  0   &  0    &  0   &  0   &  0   \\
CuPy              &  0    &  0   &  0.25  &  0   &  0   &  0   \\
pyCUDA            &  0    &  0   &  0    &  0   &  0   &  0   \\
Numba             &  0    &  0   &  0    &  0   &  0   &  0   \\
\hline
Prefix <kernel>   &       &       &      &       &       &    \\
Post fix def    &       &       &      &       &       &    \\
numpy             &  0.75  &  0.25  &  0.25 &  0.5  &  0.5  &  0.75  \\
CuPy              &  0.5   &  0.25  &  0.25 &  0.25  &  0.25  &  0.25  \\
pyCUDA            &  0.5   &  0.25  &  0.5  &  0.5  &  0.25  &  0    \\
Numba             &  0.25  &  0    &  0   &  0  &  0   &  0    \\
\hline
\hline
\end{tabular}
\caption{Metric assessment for Copilot's outputs using the input prompt pattern \texttt{<kernel> <programming model> (def)} for Python.}
\label{tab:pythonresults}
\end{center}
\end{table}

\begin{figure}[h]
\begin{center}
\includegraphics[width=0.45\textwidth]{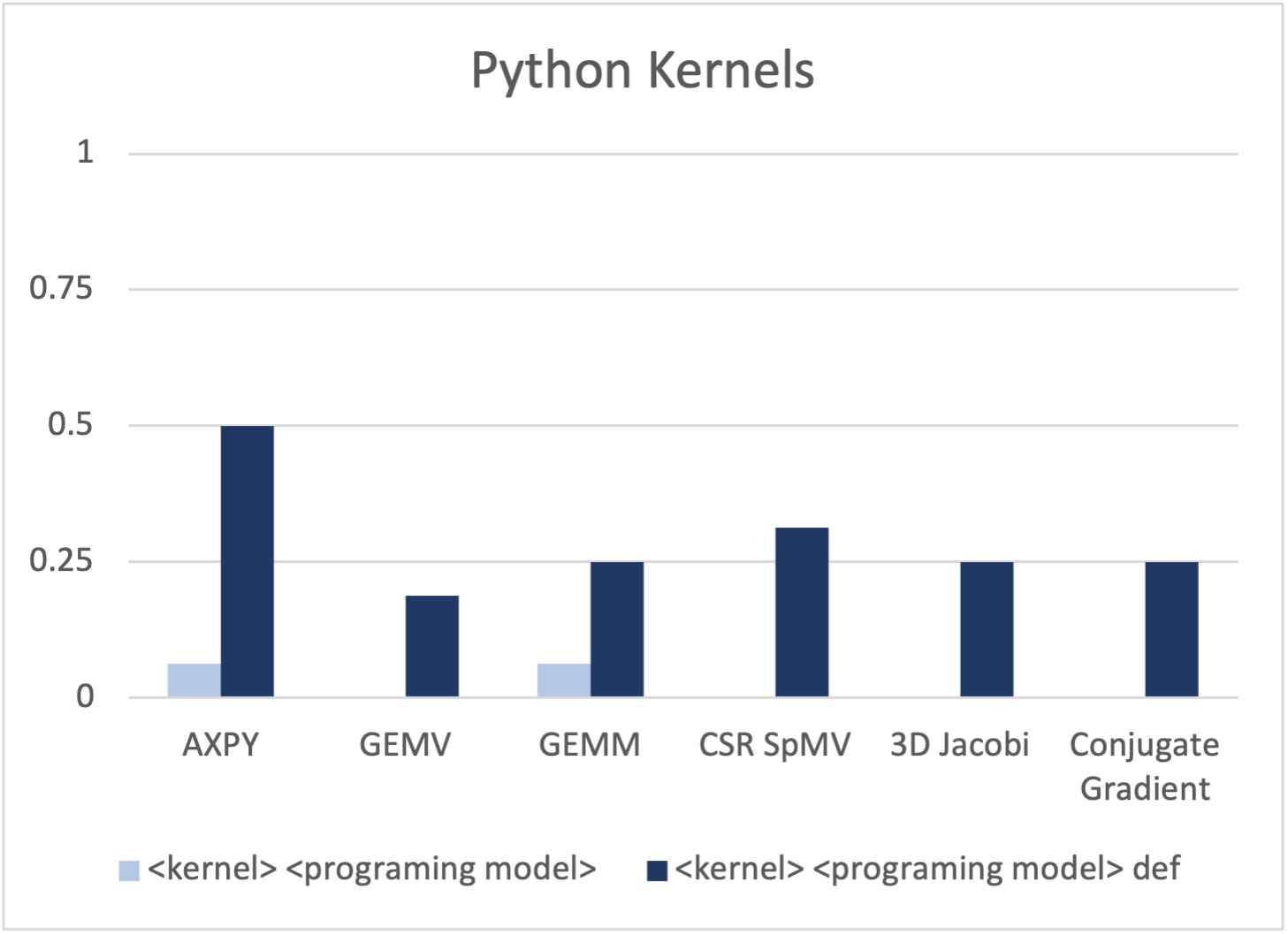}
\includegraphics[width=0.45\textwidth]{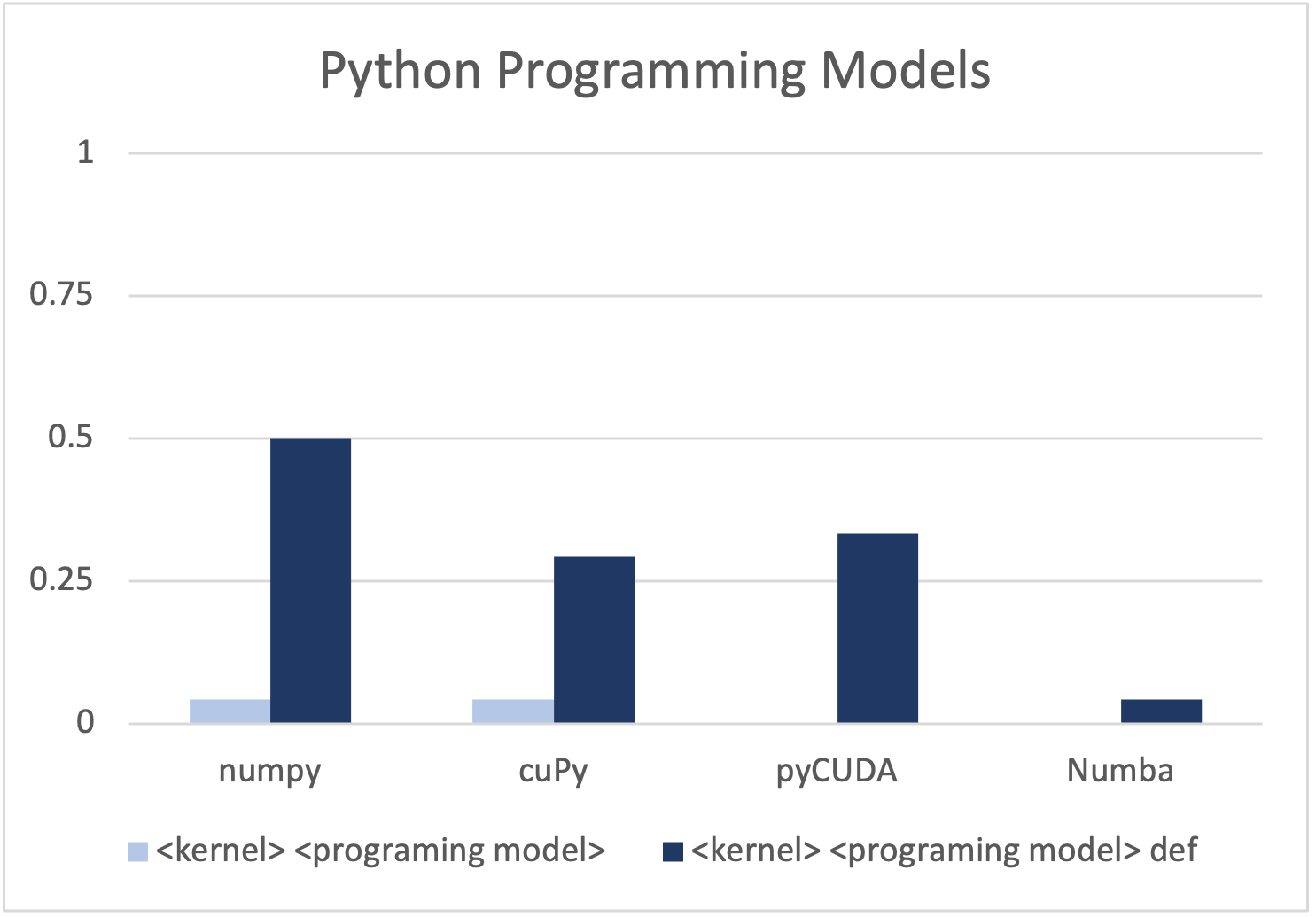}
\caption{Results for Python kernels (top) and programming models (bottom).}
\label{fig:pythonresults}
\end{center}
\end{figure}

As seen in Figure~\ref{fig:pythonresults}, the resulting kernels can have varying degrees of success, but most return at least one correct answer. This is perhaps attributed to the wide availability of Python and numpy code in public repositories. The trend also confirms the lack of correct Numba results, and perhaps it is an opportunity to exploit its pure-Python nature. Although, as recently highlighted by Kailasa et al.~\cite{kailasa2023pyexafmm}, writing Numba code for complex algorithms can be as challenging as using a compiled language.

\subsection{Julia}

We included the Julia language in our experiments because of its mathematical and performance focus. Julia provides an interesting proposition for building a dynamic language on top of LLVM and is heavily influenced by Fortran in its syntax for targeting scientific computing problems. Julia provides an accessible programming model for CPU Threads, which is part of the base language, for vendor-specific CUDA.jl and AMDGPU.jl, and for KernelAbstraction.jl for portable kernels across vendors. In previous work~\cite{godoy2023evaluating}, we showed promising results, albeit with some gaps, in a simple performance comparison on CPU and GPU runs. 

\begin{figure}[!ht]
\begin{center}
\includegraphics[width=0.45\textwidth]{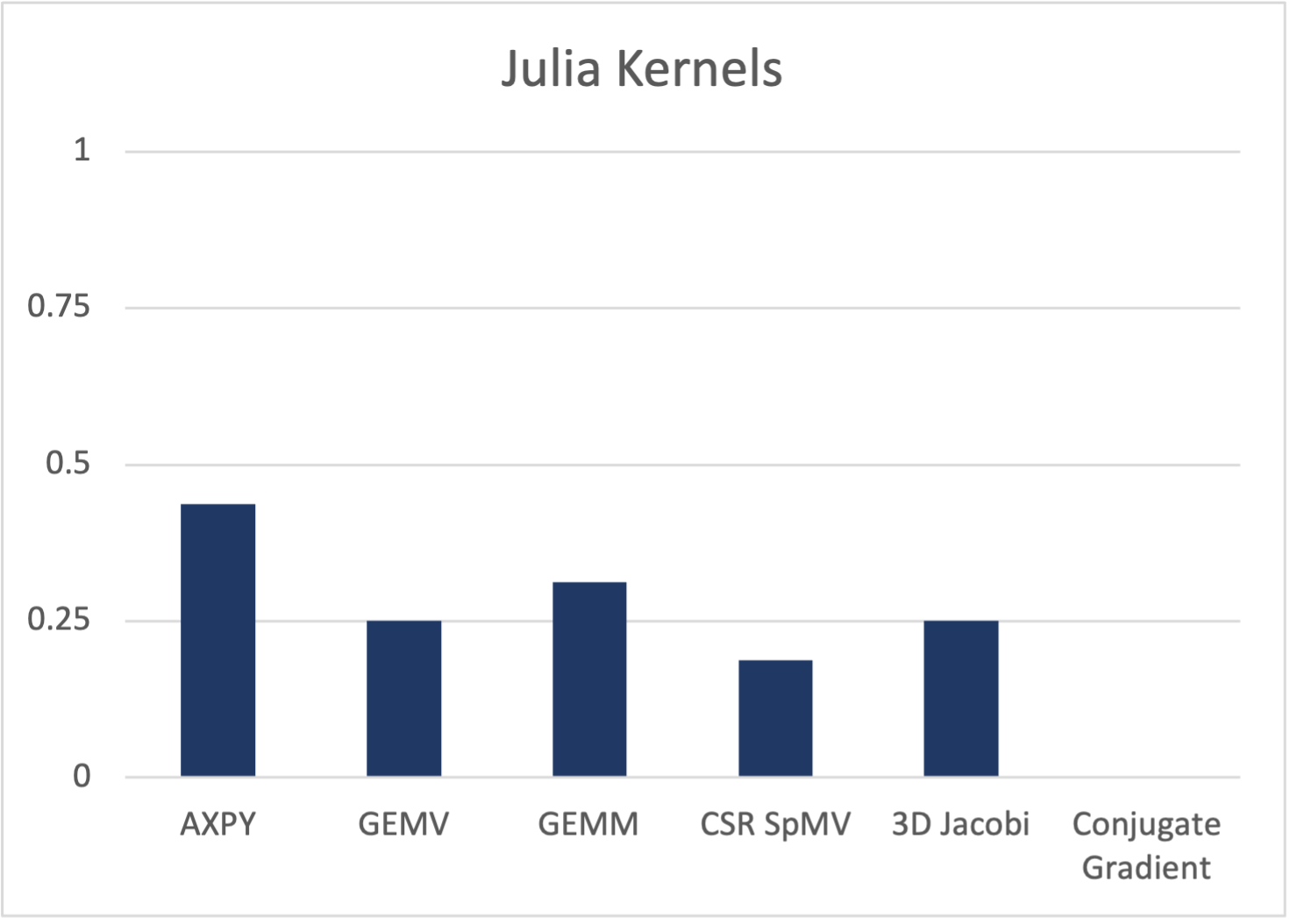}
\includegraphics[width=0.45\textwidth]{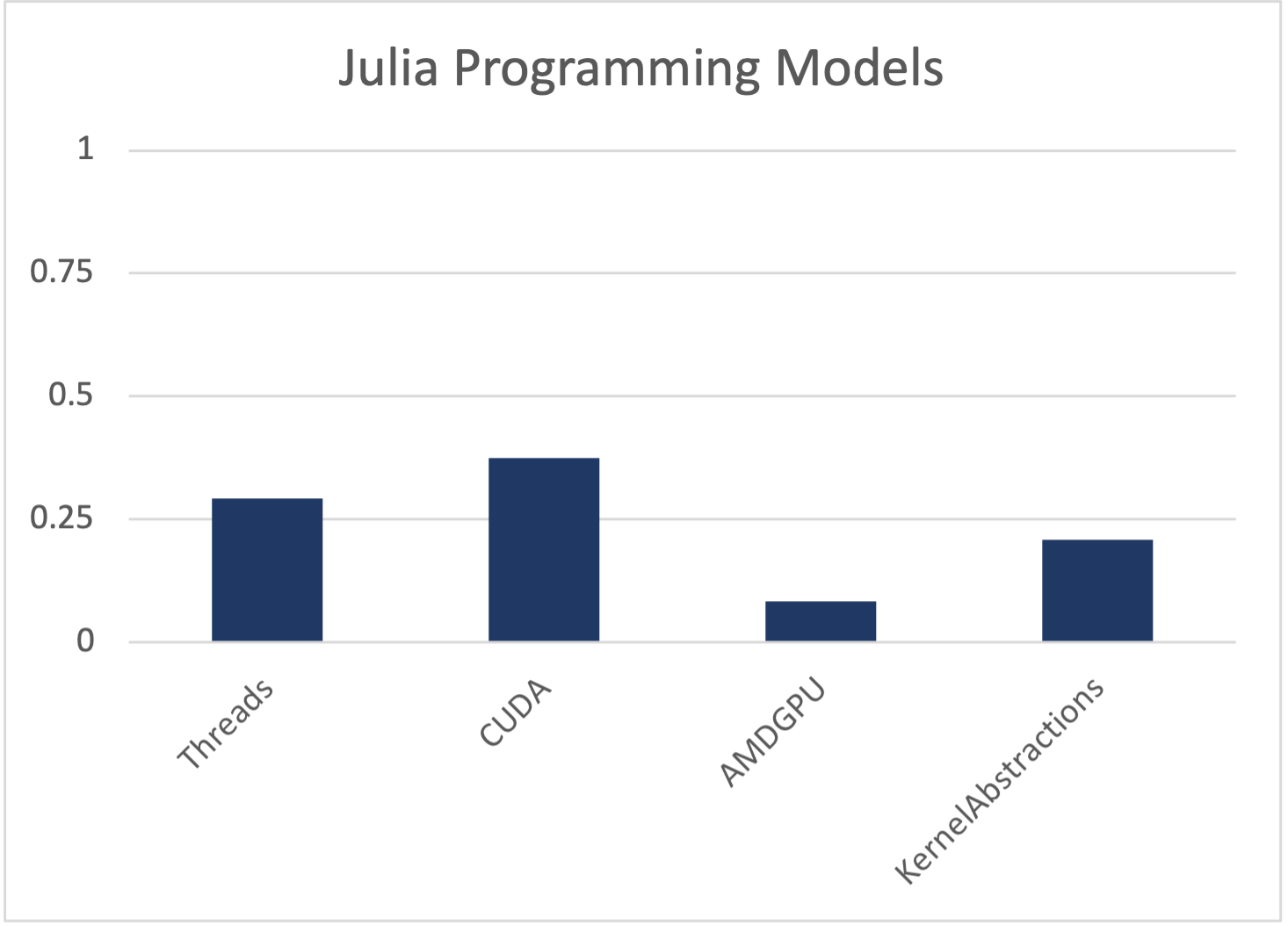}
\caption{Results for Julia kernels (top) and programming models (bottom).}
\label{fig:juliaresults}
\end{center}
\end{figure}

Table~\ref{tab:juliaresults} and Figure~\ref{fig:juliaresults} show the results for each kernel and programming model in our evaluation. 
For each programming model, we see a proficiency level between {\it novice} and {\it learner} for Threads (part of the base language) and CUDA.jl, which are the most used and mature programming models. AMDGPU.jl and KernelAbstractions.jl rank lower, which correlates with their novelty and the more limited availability of example code that targets non-NVIDIA GPU hardware.

Like with C\texttt{++}, a more complex kernel led to fewer correct results. For example, we could not get an appropriate implementation for the CG kernel (a multikernel algorithm). Unsurprisingly, as in the case of Fortran, Julia's mathematical nature allowed the OpenAI Codex to find and suggest appropriate solutions in other kernels, despite Julia being a relatively new language with fewer publicly available codes, especially compared to C\texttt{++} and Python. Notably, our results did not necessarily improve by adding more information to the prompt (e.g., the \texttt{function} language keyword). This could be an advantage for domain-specific syntax, as in the case of Fortran, because most existing codes have a very targeted use with fewer words.

\begin{table}[!h]
\begin{center}
\begin{tabular}{lllllll}
 Prompt         &  \rot{AXPY} & \rot{GEMV} & \rot{GEMM} & \rot{SpMV} & \rot{Jacobi} & \rot{CG} \\        
 \hline
Prefix <kernel>    &       &      &       &      &      &      \\
Threads            & 0.75 & 0.25  & 0.5  & 0 & 0  &  0   \\
CUDA               & 0.75 & 0.5   & 0.5  & 0.25  & 0.25  &  0   \\
AMDGPU             &  0  &  0   &  0  &  0.25   &  0   &  0   \\
KernelAbstractions & 0.25 & 0.25  & 0.25 & 0.25  & 0.25  &  0   \\
\hline
\hline
\end{tabular}
\caption{Metric assessment for Copilot's outputs using the input prompt pattern \texttt{<kernel> <programming model>} for Julia.}
\label{tab:juliaresults}
\end{center}
\end{table}

\subsection{Discussion}

\begin{figure}[!ht]
\begin{center}
\includegraphics[width=0.45\textwidth]{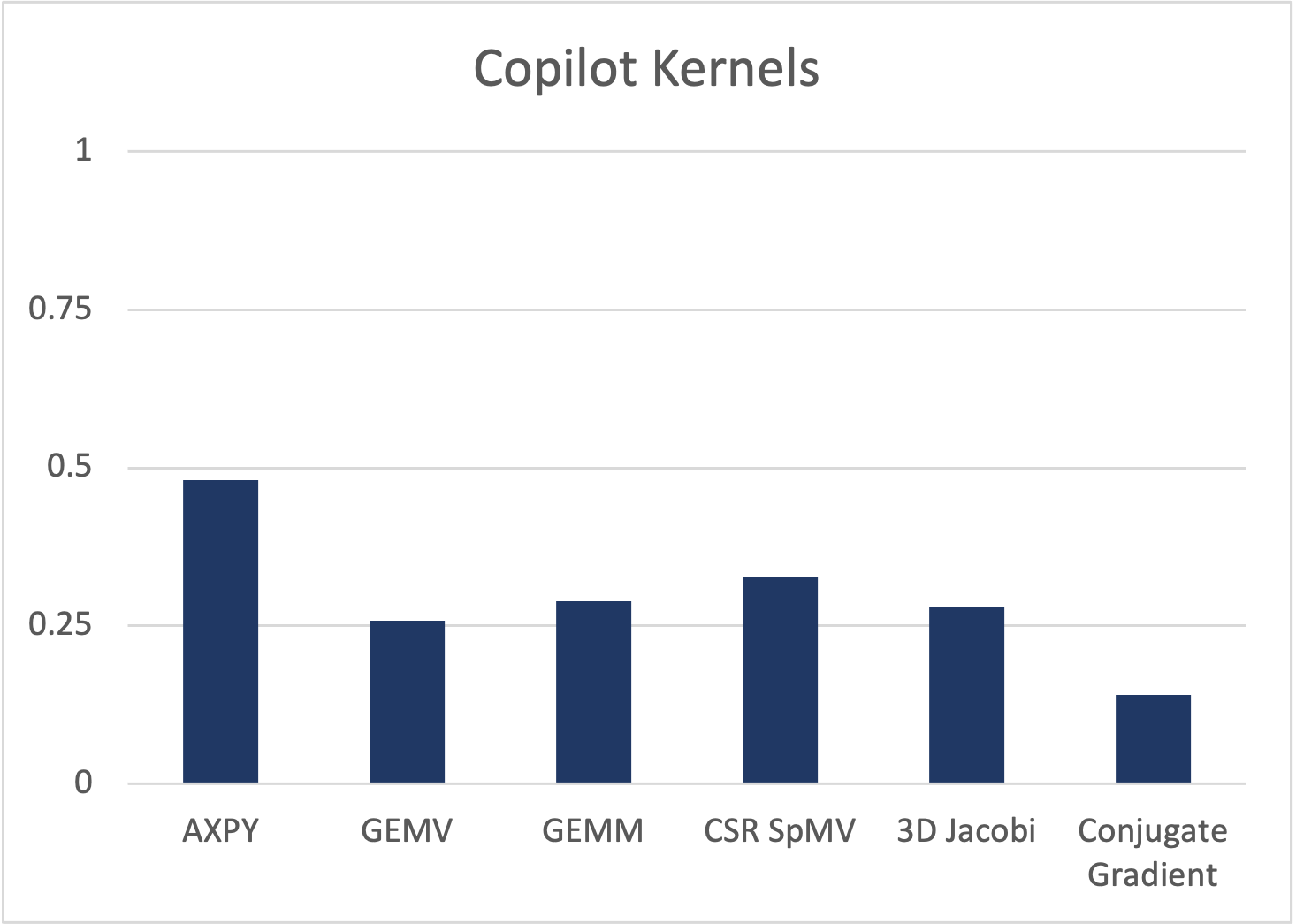}
\includegraphics[width=0.45\textwidth]{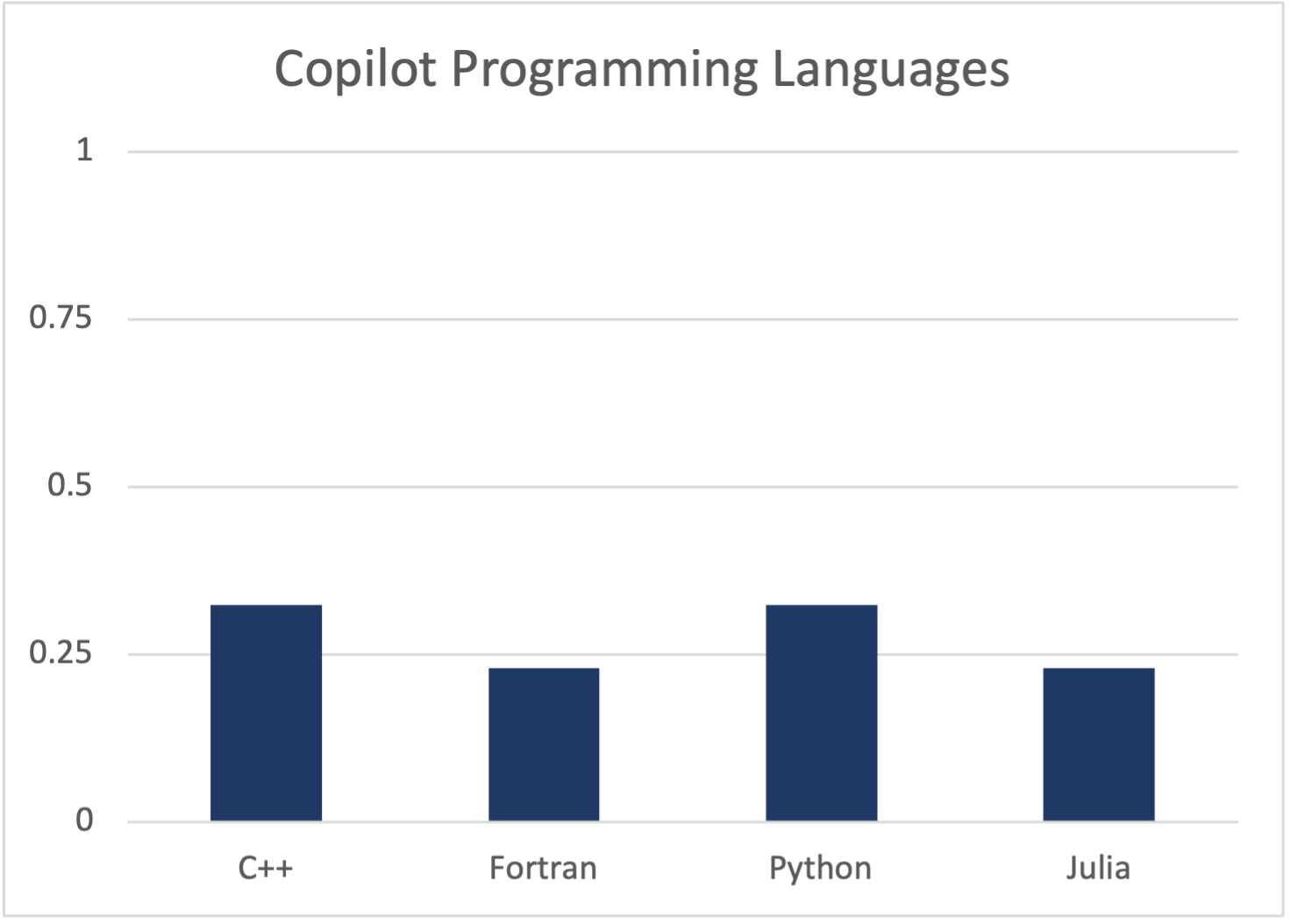}
\caption{Overall results for Copilot kernels (top) and programming languages (bottom).}
\label{fig:copilotresults}
\end{center}
\end{figure}

To gain insight into the performance of OpenAI Codex across different languages and kernels, we collated the results in Figure~\ref{fig:copilotresults}. As depicted in the graph, the complexity of the kernel directly impacts the quality of the results obtained. In other words, it becomes increasingly challenging to achieve acceptable results as the kernel's complexity increases. 

We see a relatively low level of proficiency in OpenAI Codex with an average of {\it novice} level 
for the languages and kernels tested. We see a slightly higher quality for the more popular C\texttt{++} and Python languages than for Julia or Fortran. This may be related to the popularity, maturity, and accessibility of public codes implemented in those languages. However, Fortran and Julia provide very acceptable results. The results confirm that the question is more about targeted quality than quantity in a specific domain. 

We summarize our results as follows:
\vspace{-0.5cm}
\begin{itemize}
\item As code complexity increases, obtaining acceptable results becomes more challenging.

\item Generating high-quality multistep or multikernel codes (e.g., CG) can be difficult.

\item Using keywords can improve the proficiency of the answers, but it is essential to choose the correct words that are specific and sensitive to each programming language/model or community.

\item Although the popularity or accessibility of a programming language or public code can be important, less popular languages can also provide good results owing to their targeted nature.
\end{itemize}

\section{Conclusions and Future Directions}
\label{sec:Conclusions}

We carried out an initial study to evaluate the current capacity of OpenAI Codex via Copilot for the generation of HPC numerical kernels that target parallel programming models in C\texttt{++}, Fortran, Python, and Julia. Despite current limitations, we believe that generative AI can have an extraordinary and beneficial impact on HPC software development, maintenance, and education in the future.

The research community still has several gaps to understand, and one such gap is the need for a more comprehensive taxonomy, akin to natural language, to evaluate accuracy and trustworthiness. Although our proposed taxonomy was beneficial for this initial study, it is imperative that such metrics be expanded to create a widely accepted methodology that the entire community can use. Therefore, a standard and recognized approach is needed to ensure uniformity when evaluating results across studies.

The emergence of LLM technologies such as GPT-3 and other generative AI tools presents significant questions about their integration into the future HPC software development ecosystem. For instance, can a human-in-the-loop and compiler be incorporated to refine initial LLM suggestions? Can metadata-rich suggestions be incorporated to facilitate a human decision-making process? Additionally, how do significant HPC software modernization initiatives similar to DARPA's High Productivity Computing Systems \cite{DONGARRA20081} or the US Department of Energy's Exascale Computing Project \cite{doi:10.1177/1094342010391989,8528398} incorporate these novel tools?

The automation of ecosystem features, such as building systems, packaging, validation \& verification, reproducibility, and continuous integration/continuous deployment pipelines, could significantly impact the HPC community. The technologies that put today's imperfect information closer to the human in question could redefine the educational aspects of HPC. Therefore, it is crucial to understand how each community can leverage these revolutionary capabilities to further advance their respective domains.

\section*{Acknowledgment}
This work is funded by Bluestone, an X-Stack project in the DOE Advanced Scientific Computing Office with program manager Hal Finkel.

\bibliographystyle{ACM-Reference-Format}
\bibliography{references-gpt,gpu_reference,references,vetter}

\appendix

\section{Artifact Description}
\label{ap1:Artifact}

The entire prompt input and resulting output sets are publicly available at: \url{https://github.com/keitaTN/Copilot-hpc-kernels}.
Due to the rapid evolution and statistical nature of these technologies reproducibility and replicability of the present results is a challenging aspect that we expect to improve over time, but has no guarantees as of today. Hence, it's imperative that the reproducibility information is provided.

Some additional information: 

\begin{itemize}
    \item Experiments used the Visual Studio Code GitHub Copilot plugin service on three separate Linux systems using Ubuntu 22.04
    \item Experiments were carried between April 14th and April 21st of 2023.
    \item The raw dataset are identified by \\ \texttt{<kernel>/<language>/<kernel>\_outputs.<ext>} \\ directory and file structure. 
    \item Each Prompt used in this study is identified with the line comment \\ \texttt{ Prompt: <kernel> <programming model> <additional text>}.
\end{itemize}


\end{document}